\documentclass[letterpaper]{article} 
\usepackage{aaai2026}  
\usepackage{times}  
\usepackage{helvet}  
\usepackage{courier}  
\usepackage[hyphens]{url}  
\usepackage{graphicx} 
\usepackage{natbib}  
\usepackage{caption} 
\usepackage{algorithm}
\usepackage{algorithmic}
\usepackage{booktabs}
\usepackage{makecell}
\usepackage{multirow}
\usepackage{subcaption}
\usepackage{xcolor}
\usepackage{amsfonts}
\usepackage{fvextra}
\usepackage{newfloat}
\usepackage{listings}
\usepackage{amsmath}
\DeclareCaptionStyle{ruled}{labelfont=normalfont,labelsep=colon,strut=off} 
\floatstyle{ruled}
\newfloat{listing}{tb}{lst}{}
\floatname{listing}{Listing}
\title{LTA-thinker: Latent Thought-Augmented Training Framework for Large Language Models on Complex Reasoning}
\author{
    Jiaqi Wang\textsuperscript{\rm 1},
    Binquan Ji\textsuperscript{\rm 2},
    Haibo Luo\textsuperscript{\rm 2},
    Yiyang Qi\textsuperscript{\rm 2},
    Ruiting Li\textsuperscript{\rm 2},
    Huiyan Wang\textsuperscript{\rm 2},
    Yuantao Han\textsuperscript{\rm 3},
    Cangyi Yang\textsuperscript{\rm 3},
    Jiaxu Zhang\textsuperscript{\rm 3},
    Feiliang Ren\textsuperscript{\rm 3}\thanks{\raggedright Corresponding author.}
}
\affiliations{
    \textsuperscript{\rm 1}School of Computer Science and Engineering, Northeastern University\\
    195 Chuangxin Road, Hunnan District\\
    Shenyang, Liaoning 110169, China\\
    \textup{2310742@stu.neu.edu.cn}, \textup{renfeiliang@cse.neu.edu.cn}

}

\usepackage{bibentry}

\begin{document}

\maketitle

\begin{abstract}
Complex Reasoning in Large Language Models can be dynamically optimized using Test-Time Scaling (TTS) to mitigate Overthinking. Methods such as Coconut, SoftCoT and its variant are effective in continuous latent space inference, the core bottleneck still lies in the efficient generation and utilization of high-quality Latent Thought. Drawing from the theory of SoftCoT++ that a larger variance in the generated Latent Thought distribution more closely approximates the golden truth distribution, we propose a Latent Thought-Augmented Training Framework--LTA-Thinker, which improves distributional variance and enhances reasoning performance from two perspectives. First, LTA-Thinker constructs a Latent Thought generation architecture based on a learnable prior. This architecture aims to increase the variance distribution of generated Latent Thought Vectors in order to simplify the overall structure and raise the performance ceiling. Second, LTA-Thinker introduces a distribution-based directional optimization paradigm that jointly constrains both distribution locality and distribution scale. This mechanism improves information efficiency and computational cost through a multi-objective co-training strategy, which combines standard Supervised Fine-Tuning (SFT) loss with two novel losses: Semantic Alignment Loss, which utilizes KL divergence to ensure that the Latent Thought is highly relevant to the semantics of the question; Reasoning Focus Loss, which utilizes a contrastive learning mechanism to guide the model to focus on the most critical reasoning steps. Experiments show that LTA-thinker achieves state-of-the-art (SOTA) performance among various baselines and demonstrates a higher performance ceiling and better scaling effects.
\footnote{
Code at: \url{https://github.com/wangjiaqi886/LTA-Thinker}}
\end{abstract}


\section{Introduction}

In recent years, Large Language Models (LLMs)~\cite{yang2025qwen3technicalreport,deepseekai2025deepseekr1incentivizingreasoningcapability,geminiteam2025geminifamilyhighlycapable,DBLP:journals/corr/abs-2303-08774,DBLP:journals/corr/abs-2407-21783,DBLP:journals/corr/abs-2406-12793,DBLP:journals/corr/abs-2501-12599} have achieved revolutionary progress on various natural language processing tasks. With the emergence of large reasoning models, LLMs have demonstrated outstanding performance on tasks requiring multi-step, Complex Reasoning, such as mathematical problem solving, code generation, and strategic planning. However, they still face the "Overthinking" problem. This issue causes the reasoning process to produce verbose, inefficient, or off-topic output. Test-Time Scaling (TTS)~\cite{DBLP:journals/corr/abs-2503-24235} mitigates the issue of model performance being negatively affected by Overthinking~\cite{DBLP:journals/corr/abs-2503-16419} in the output by employing methods of parallel scaling, sequential scaling, or hybrid scaling during the reasoning process.

\begin{figure}[t]
\centering
\includegraphics[width=0.95\columnwidth]{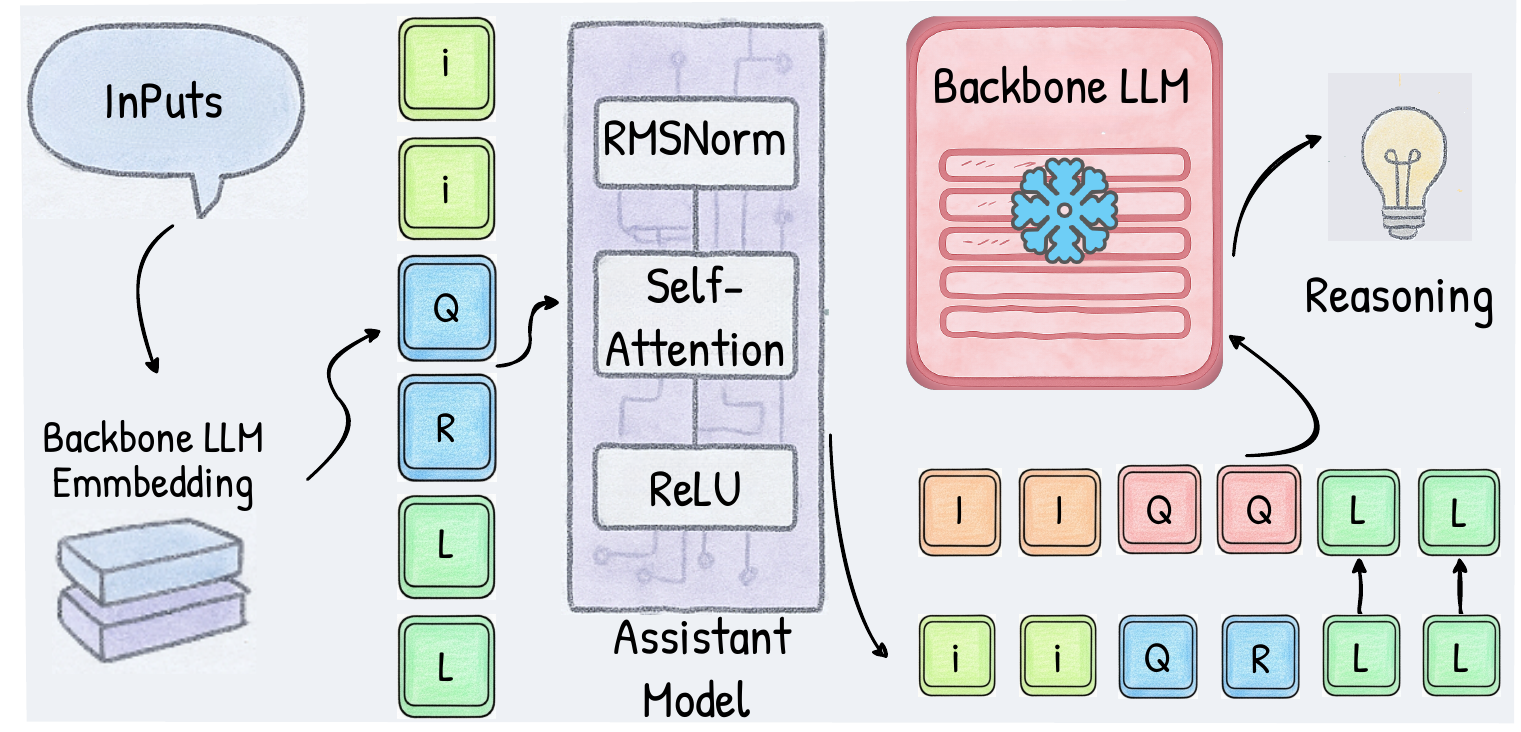} 
\caption{Process diagram of LTA-Thinker “thinking” in a continuous latent space. In the figure, "I" and "i" denote Instruction, "Q" denotes question, "R" denotes reasoning steps, and "L" denotes Latent Thought.}
\label{fig1}
\end{figure}

 Most Test-Time Scaling methods improve performance by influencing outputs in the discrete token space~\cite{DBLP:journals/corr/abs-2502-05171,DBLP:journals/corr/abs-2405-14838}. However, reasoning in a discrete space can easily overlook more detailed or critical information. To overcome the limitations of discrete reasoning, researchers have begun to explore a paradigm of "thinking" in a continuous Latent Space. Such as Coconut~\cite{DBLP:journals/corr/abs-2412-06769}, CODI~\cite{DBLP:journals/corr/abs-2502-21074}, CCOT~\cite{DBLP:journals/corr/abs-2412-13171},  SoftCoT~\cite{DBLP:conf/acl/00010ZM25} and its variants~\cite{DBLP:journals/corr/abs-2505-11484}. These methods generate continuous Latent Thought Vectors to guide reasoning. They avoid generating lengthy text and provide flexible hidden state representations. However, these approaches still face challenges in theory and practice, including insufficient information utilization, structural redundancy, inefficient computation, and suboptimal performance. For instance, Coconut first proposes the method of "thinking" in a continuous Latent Space, but its performance is poor. SoftCoT utilizes a trained small LLM as an auxiliary model to generate Latent Thought. However, due to low information utilization, its performance is poor. SoftCoT++, a variant of SoftCoT, theoretically proves that under restricted conditions, the larger the variance of the distribution of generating Latent Thought, the closer the distribution is to the golden truth distribution. However, this method directly takes maximizing variance as its optimization objective, leading to the Latent Thought being dominated by excessively large, uninformative variance, while also significantly increasing training costs and causing model structure redundancy.

To address the aforementioned challenges, we propose a Latent Thought-Augmented Training Framework for Large Language Models on Complex Reasoning--LTA-Thinker. The process diagram of LTA-Thinker “thinking” in a continuous latent space is shown in Figure \ref{fig1}. A lightweight assistant model takes the "Inputs" initialized from the embedding layer of the Backbone LLM and generates Latent Thought Vectors. The input of the backbone LLM is augmented with these Latent Thoughts, thereby guiding it toward more accurate and efficient reasoning. LTA-Thinker follows the lemmas and assumptions from SoftCoT++. Its core idea is to optimize variance and approximate the golden truth distribution from two perspectives. The first increases the upper bound of the variance of the Latent Thought generation distribution through theoretical proof and model design. The second introduces a distribution-based directional optimization paradigm to optimize the "direction" and "shape" of the Latent Thought generation distribution in terms of Distribution Locality and Distribution Scale.

Specifically, LTA-Thinker continues the paradigm of "thinking" in a continuous Latent Space. It injects optimized continuous Latent Thought Vectors into the LLM’s input. These vectors, called "Soft Thoughts", guide the LLM to generate more precise and efficient reasoning paths. The first perspective uses a lightweight thought generation module, which is responsible for encoding the original question into a fixed number of Latent Thought vectors. During inference, these vectors replace preset placeholders in the input sequence of the main LLM to guide it in generating the chain of thought and the final answer. The second perspective employs a multi-objective joint loss function for directional optimization. This loss function consists of three conponents. The first is a distributional foundation constraint, the standard cross-entropy loss. It ensures acurate and coherent output. The second is a distribution locality constraint, a semantic alignment loss. This loss minimizes the KL divergence between the Latent Thought distribution and the core semantic representation distribution of the question. The third is a distribution scale constraint, a reasoning focus loss. It utilizes contrastive learning to expand the variance of the Latent Thought distribution. This helps enabling the model to focus on critical reasoning steps. The Backbone LLM parameters in LTA-Thinker remain frozen.

The innovations of LTA-Thinker are as follows:
\begin{itemize}
\item \textbf{Learnable Prior-Based Latent Thought Generation Architecture: }It discards the traditional pretrained small LLM and adopts a more lightweight module to generate Latent Thought vectors. This expands the variance space of the Latent Thought distribution, providing greater flexibility for subsequent optimization.
\item \textbf{Distribution-based Directional Optimization Para-digm: }High-variance random distributions without structural information are meaningless. Thus, in addition to the standard supervised fine-tuning (SFT) loss, we introduce two auxiliary loss functions that together form a multi-objective collaborative training strategy:
\begin{itemize}
 \item \textbf{Semantic Alignment Loss (Position Constraint):} It minimizes the KL divergence between the Latent Thought vectors and the question representation. This anchors the center of the Latent Thought distribution to the core semantics of the question.
 \item \textbf{Reasoning Focus Loss (Scale Constraint): }It utilizes a contrastive learning mechanism to focus on and enhance the variance of the Latent Thought distribution. In addition, it guides the model to focus on the most critical reasoning step.
\end{itemize}
\item The LTA-Thinker framework has achieved SOTA performance in multiple Complex Reasoning benchmarks. This validats the effectiveness of the Latent Thought generation architecture. It also confirms the effectiveness of the distribution-based directional optimization paradigm.

\end{itemize}

\section{Related Work}
The reasoning capabilities of Large Language Models (LLMs) can be enhanced through various computational scaling strategies. According to a survey\cite{DBLP:journals/corr/abs-2503-24235}, these strategies can be broadly categorized into parallel, sequential, and hybrid methods, which differ in how they utilize computational resources to explore the solution space. Considering that our work introduces information from reasoning steps, this paper focuses on sequential methods. (1) Parallel scaling \cite{Stanovich_West_2000,DBLP:journals/corr/abs-2502-14382,DBLP:journals/corr/abs-2407-21787,DBLP:conf/emnlp/Renze24,DBLP:journals/corr/abs-2411-15124,DBLP:conf/iclr/0002WSLCNCZ23}involves generating multiple outputs in parallel and then aggregating them into a single answer. (2) Sequential scaling introduces a series of intermediate steps, guiding the model to think step-by-step and iteratively refine its solution to arrive at a final answer. This approach takes various forms, including chain-of-thought, which guides the model's step-by-step thinking \cite{DBLP:conf/nips/Wei0SBIXCLZ22}, refining responses \cite{DBLP:conf/nips/MadaanTGHGW0DPY23}, and decomposing complex questions into sub-questions to be solved one by one \cite{DBLP:conf/iclr/ZhouSHWS0SCBLC23,DBLP:conf/nips/ZelikmanWMG22}. Research \cite{DBLP:conf/iclr/ChenLSZ24,DBLP:conf/iclr/GouSGSYDC24,DBLP:conf/iclr/Snell0XK25,DBLP:journals/corr/abs-2502-05449} indicates that this process of iterative correction can trigger the self-correction abilities of the model, leading to better performance on complex tasks. (3) Hybrid scaling methods \cite{DBLP:journals/corr/abs-2502-05449,DBLP:conf/iclr/Snell0XK25,DBLP:conf/nips/YaoYZS00N23,DBLP:conf/aaai/BestaBKGPGGLNNH24,DBLP:journals/corr/abs-2502-11169,DBLP:conf/iclr/WangWAZZ25,DBLP:journals/corr/abs-2503-04625} complementarily leverage the advantages of parallel and sequential scaling.

Latent Thought reasoning methods can be broadly divided into two categories. The first involves fine-tuning an LLM to directly utilize latent representations for reasoning. The second uses auxiliary modules to enhance a frozen LLM. In the first category, Coconut \cite{DBLP:journals/corr/abs-2412-06769} pioneers the use of the last layer hidden states of the model as "continuous thought" to replace discrete text tokens; CODI \cite{DBLP:journals/corr/abs-2502-21074} employs self-distillation techniques, which enables internal reasoning without generating explicit CoT. CCOT \cite{DBLP:journals/corr/abs-2412-13171} trains a compression module to condense lengthy reasoning steps into crucial "meditation" tokens. The second category focuses on enabling Latent Thought reasoning without altering the LLM's parameters. A representative work in this area is SoftCoT \cite{DBLP:conf/acl/00010ZM25}, which uses a lightweight auxiliary model to generate specific "soft thought" tokens and projects them into the embedding space of a frozen LLM. SoftCoT++ \cite{DBLP:journals/corr/abs-2505-11484}, a variant of SoftCoT, theoretically proves the relationship between variance and distribution in Latent Thought reasoning and leverages this theory to maximize the variance among Latent Thoughts to achieve optimal performance. Overall, these works collectively point to an important research direction: how to efficiently generate and utilize high-quality Latent Thought to find a better balance between reasoning depth and computational efficiency.

\begin{figure*}[t]
\centering
\includegraphics[width=0.95\textwidth]{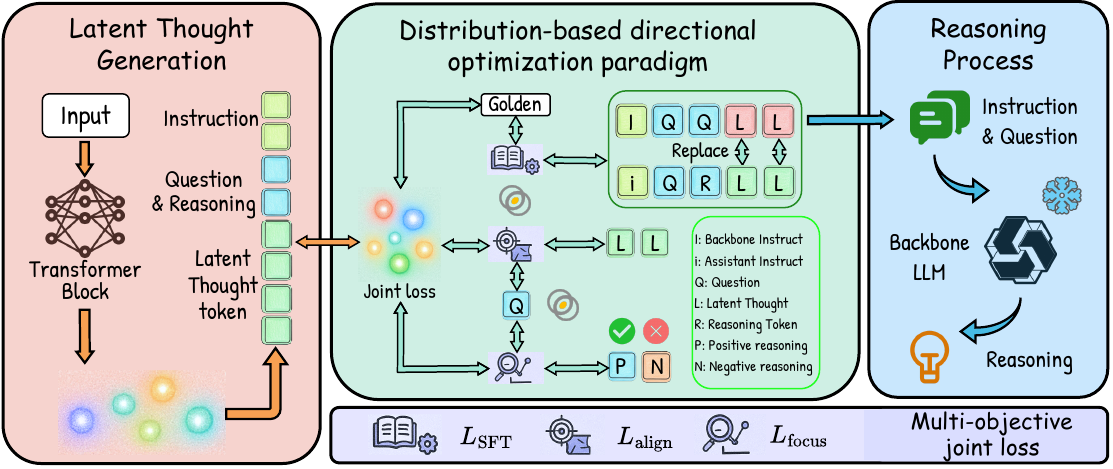} 
\caption{The LTA-Thinker overall training framework is composed of three parts. First, Latent Thought Generation, depicts the process where Latent Thought is generated through a Transformer Block, with the input initialized by the embedding layer of a Backbone Large Language Model (LLM). Second, the Distribution-based directional optimization paradigm, illustrates the Latent Thought Token replacement process and the roles of semantic alignment loss and reasoning focusing loss within the three designated losses. Third, reasoning Process, the backbone LLM conducts reasoning and generates responses upon receiving the input augmented with the Latent Thought.}
\label{fig2}
\end{figure*}

\section{Methodology}
\subsection{Problem Definition}

To introduce the Latent Thought Tokens and extend it to the problem-solving process of Large Language Models (LLMs), we formally decompose the entire reasoning process into three continuous auto-regressive stages. The process begins with a task instruction $I$ and an input question $ Q $, and proceeds sequentially through the following stages:
\subsubsection{Thinking Stage:}
The model generates a series of intermediate thought steps $ L = \{l_1, l_2, \dots, l_n\} $ based on $ I $ and $ Q $, where each $ l_i $ represents a Latent Thought token. This process is modeled as a conditional probability:
\begin{equation}
P(L \mid I, Q) = \prod_{i=1}^{n} P(l_i \mid I, Q, l_{<i})
\end{equation}
where $ l_{<i} $ denotes the previously generated thought steps.
\subsubsection{Reasoning Stage:}
Based on $ L $, the model generates an explicit reasoning process $ R = \{r_1, r_2, \dots, r_m\} $, which is a readable explanation of the problem solving logic. Its generation process can be formulated as follows:
\begin{equation}
P(R \mid I, Q, L) = \prod_{j=1}^{m} P(r_j \mid I, Q, L, r_{<j})
\end{equation}
where $ r_{<j} $ denotes the reasoning content already generated.
\subsubsection{Answer Generation Stage:}
The model synthesizes all the information to generate the answer $ A = \{a_1, a_2, \dots, a_k\} $, with its probability being:
\begin{equation}
P(A \mid I, Q, L, R) = \prod_{k=1}^{o} P(a_k \mid I, Q, T, R, a_{<k})
\end{equation}
where $ a_{<k} $ denotes the generated answer part.

The joint probability of the entire generation process can be decomposed into the following formula:
\begin{equation}
\begin{split}
P(L, R, A \mid I, Q) &= P(L \mid I, Q) \cdot P(R \mid I, Q, L) \\
                     &\quad \cdot P(A \mid I, Q, L, R)
\end{split}
\end{equation}
\subsection{Assumptions and Lemmas}
For the Thinking Stage mentioned above, inputs of the model are  $ I $ and $ Q $, and its output is a series of continuous Latent Thought tokens. Based on this, we introduce the assumptions and lemmas from SoftCoT++\cite{DBLP:journals/corr/abs-2505-11484} for a more refined modeling:
\subsubsection{Assumption:}
For a given task instruction $ I $ and input question $ Q $, there exists a smooth and differentiable distribution $ P_{\text{real}}(L \mid I, Q) $, such that the deterministically generated Latent Thought representation $ L_{\text{Latent}} $ can be considered a sample from this distribution: $ L_{\text{Latent}} \sim P_{\text{real}}(L \mid I, Q) $
\subsubsection{Lemma 1:}
If the magnitude of the perturbation vector $ \delta $ is sufficiently small, then the perturbed sample $ L_{\text{Latent}} + \delta $ will still be within a high-probability density region of $ P_{\text{real}} $. The proof of Lemma 1 is given in the Appendix.

\subsubsection{Definition 1:}
Given a set of small perturbations $ \{\delta_i\}_{i=1}^n $
, the perturbed Latent Thought token is defined as $ L_p^i = L_{\text{Latent}} + \delta_i $. The empirical distribution $ Q_1 $ is estimated from the set of perturbed samples $ \{L_p^i\}_{i=1}^n $
\begin{equation}
Q_1 = \text{EmpiricalDist}(\{L_p^i\}_{i=1}^n)
\end{equation}
\subsubsection{Definition 2:}
Let $ \{L_{\text{Latent}}^i\}_{i=1}^n $ be a set of samples drawn directly from the golden truth distribution $ P_{\text{real}} $. The empirical distribution $ Q_2 $ is estimated from $ \{L_{\text{Latent}}^i\}_{i=1}^n $:
\begin{equation}
Q_2 = \text{EmpiricalDist}(\{L_{\text{Latent}}^i\}_{i=1}^n )
\end{equation}
\subsubsection{Lemma 2:}
Assume that the variance of the golden truth distribution $ P_{\text{real}} $ is $ \text{Var}[P_{\text{real}}]>0 $, and the magnitude of the perturbation is $ \delta_i < \text{Var}[P_{\text{real}}] $. If the variances of the empirical distributions $ Q_1 $ and $ Q_2 $ satisfy:
\begin{equation}
\text{Var}[Q_1] < \text{Var}[Q_2] \leq \text{Var}[P_{\text{real}}]
\end{equation}
then $ Q_2 $ has a smaller KL divergence, which denotes $ \text{KL}(P \| Q_2) < \text{KL}(P \| Q_1) $. This indicates that $ Q_2 $ is closer to the golden truth distribution $ P_{\text{real}} $. The proof of Lemma 2 is given in the Appendix.


In summary, $ Q_1 $ represents generating diverse Latent Thought through small perturbations of a single sample; $ Q_2 $ represents generating diverse Latent Thought by estimating from samples of the golden truth distribution (which is the method to be implemented). The direction we need to optimize is to make the variance of $ Q_{\text{LTA-Thinker}} $ larger than the variance of $ Q_{\text{SoftCoT++}} $, such that $ \text{KL}(P \| Q_{\text{LTA-Thinker}}) < \text{KL}(P \| Q_{\text{SoftCoT++}}) $, thereby achieving a closer approximation to the golden truth distribution $P_{\text{real}}$. The diagram of LTA-Thinker training framework is shown in Figure \ref{fig2}.

\subsection{Latent Thought Generation}
In this section, we will introduce Learnable Prior-Based Latent Thought Generation. To obtain a $ Q_{\text{LTA-Thinker}} $ that is closer to the golden truth distribution $P_{\text{real}}$, we must first consider a constraint: $\text{Var}[Q_{\text{LTA-Thinker}}] \leq \text{Var}[P_{\text{real}}]$. This implies that the Latent Thought generation architecture based on a learnable prior, which we will design, must first have a controllable variance, and second, this variance should be as large as possible. In previous approaches, a small LLM is typically used as an Assistant Model to generate Latent Thought. These small LLMs are generally trained on large amounts of text. The variance of these pretrained small LLMs will inevitably satisfy $\text{Var}[Q_1] < \text{Var}[Q_{\text{LLM}}]\leq \text{Var}[P_{\text{real}}]$. However, at the same time, due to excessive text training, their variance will be confined to a narrower space:
\begin{equation}
\text{Var}[Q_{\text{LLM-}}]< \text{Var}[Q_{\text{LLM}}]\leq \text{Var}[Q_{\text{LLM+}}]< \text{Var}[P_{\text{real}}]
\end{equation}
This seems to limit the upper bound of the distribution variance $\text{Var}[Q_{\text{LLM}}]$, which in turn limits the distance between the distribution $Q_{\text{LLM}}$ and the golden truth distribution $ P_{\text{real}} $.

To address this, we have designed a Latent Thought generation architecture based on a learnable prior. This framework is built using a lightweight Transformer Block\cite{DBLP:conf/nips/VaswaniSPUJGKP17} with randomly initialized parameters. Through training, we constrain the variance of its representation distribution $Q_{\text{TB}}$ to satisfy $\text{Var}[Q_1] < \text{Var}[Q_{\text{TB}}]\leq \text{Var}[P_{\text{real}}]$. Furthermore, the upper bound of this distribution's variance satisfies: $\text{Var}[Q_{\text{LLM+}}]< \text{Var}[Q_{\text{TB+}}]< \text{Var}[P_{\text{real}}]$.

For the proposed architecture of Latent Thought generation based on a learnable prior, we expect its variance to satisfy $\text{Var}[Q_{\text{LLM+}}]<\text{Var}[Q_{\text{TB}}] \leq\text{Var}[Q_{\text{TB+}}]< \text{Var}[P_{\text{real}}]$. This involves the two optimizations for variance mentioned earlier. The first optimization is controllability, where $\text{Var}[Q_{\text{TB}}] < \text{Var}[P_{\text{real}}]$. Therefore, our proposed lightweight Transformer Block references the architecture of some current mainstream LLMs, using RMSNorm instead of LayerNorm for normalization and integrating a self-attention mechanism and a feed-forward network with ReLU activation. The second optimization is to make the variance as large as possible. Thus, we randomly initialize this module and use the directional optimization method from the next section to further expand $\text{Var}[Q_{\text{LTA-Thinker}}]$.

The Transformer Block structure is chosen for two reasons: on one hand, its highly efficient learning capability has been widely verified; on the other hand, this structure ensures that the variance of the modeled distribution satisfies $\text{Var}[Q_{\text{TB}}] < \text{Var}[P_{\text{real}}]$ after training. In contrast, simpler model structures might lead to $ \text{Var}[P_{\text{real}}]< \text{Var}[Q_{\text{TB}}]$ due to insufficient representational capacity. We will use a linear layer to prove this in ablation experiments.

\subsection{Distribution-based Directional Optimization}
In the previous section, we theoretically demonstrated that when $\text{Var}[Q_{\text{LLM+}}]< \text{Var}[Q_{\text{TB+}}]< \text{Var}[P_{\text{real}}]$, the upper bound of $Q_{\text{TB}}$ can get closer to the golden truth distribution $P_{\text{real}}$. However, a random distribution with merely high variance is meaningless. It must be given "direction" and "shape" through refined optimization, enriching it with more information so that, while maintaining high variance, it can generate high-quality Latent Thoughts that are beneficial for downstream tasks. This means the variance should satisfy:
\begin{equation}
\begin{split}
\text{Var}[Q_{\text{LLM+}}] &< \text{Var}[Q_{\text{TB}}] \leq \text{Var}[Q_{\text{LTA-Thinker}}] \\
                           &\leq \text{Var}[Q_{\text{TB+}}] < \text{Var}[P_{\text{real}}]
\end{split}
\end{equation}
To this end, we propose a distribution-based directional optimization paradigm. This paradigm co-constrains Distribution Locality and Distribution Scale, aiming to expand the initial distribution variance generated by the Transformer Block while guiding its semantic features to converge on regions crucial for Complex Reasoning. This paradigm is realized through a joint training objective, which consists of three loss function components. The total loss $L_{\text{total}}$ is defined as a weighted sum of the three sub-losses:
\begin{equation}
L_{\text{total}} = \lambda_{\text{sft}} L_{\text{SFT}} + \lambda_{\text{align}} L_{\text{align}} + \lambda_{\text{focus}} L_{\text{focus}}
\end{equation}
where $\lambda_{\text{sft}}$,$\lambda_{\text{align}}$,and$\lambda_{\text{focus}}$ are the weight hyperparameters.
\subsubsection{Distributional Foundation Constraint}
This is the standard auto-regressive language model loss, which aims to maximize the probability of the model generating the correct reasoning steps and answer. This loss is fundamental to ensure the functionality of the entire generative system. For a given training sample, its loss function is the standard cross-entropy loss, as follows:
\begin{equation}
L_{\text{SFT}} = - \sum_{t=1}^{|Y|} \log P(y_t | y_{<t}, X_{\text{aug}})
\end{equation}
where $Y = \{y_1, \dots, y_{|Y|}\}$ is the target output containing the reasoning steps and answers. $X_{\text{aug}}$ is the input augmented with the Latent Thought. This loss ensures the model possesses fundamental reasoning and generation capabilities.
\subsubsection{Distribution Locality Constraint}
To enable the Latent Thought vectors to accurately capture and represent the core semantics of the original question, we introduce a Semantic Alignment Loss. This loss aims to minimize the distance between the probability distribution of the Latent Thought vectors and the probability distribution of the question representation. Specifically, this is achieved by minimizing the KL divergence between the probability distribution of the question representation vector $\mathbf{e}_q$ and the probability distribution of the Latent Thought vector $\mathbf{v}_i$:
\begin{equation}
L_{\text{align}} = \frac{1}{N} \sum_{i=1}^{N} D_{\text{KL}}(P(\cdot|\mathbf{e}_q) || P(\cdot|\mathbf{v}_i))
\end{equation}
where $P(\cdot|\mathbf{z}) = \text{softmax}(\mathbf{W} \mathbf{z})$, the question representation vector $\mathbf{e}_q$ is initialized using the embedding layer of the Backbone LLM, and the Latent Thought vector $\mathbf{v}_i$ is output by a Transformer Block. This loss function compels the Latent Thought vectors produced by the thought generation module to remain consistent with the essence of the question on a semantic level, thereby effectively controlling the central location of the distribution and preventing it from drifting into irrelevant semantic regions.

\begin{table*}[ht]
\centering
\setlength{\heavyrulewidth}{1.2pt}
\setlength{\tabcolsep}{4.5pt} 
\begin{tabular}{@{}llccccc@{}}
\toprule
\multirow{2.5}{*}{\makecell[b]{Base LLM}} & 
\multirow{2.5}{*}{\makecell[b]{Baselines}} & 
\multicolumn{3}{c}{Mathematical} & 
\multicolumn{1}{c}{Commonsense} & 
\multicolumn{1}{c}{Symbolic} \\
& & 
\makecell[b]{GSM8K} & 
\makecell[b]{AQuA} & 
\makecell[b]{Avg.(Math)} & 
\makecell[b]{StrategyQA} & 
\makecell[b]{DU} \\ 
\midrule

\multirow{5}{*}{\makecell[l]{Qwen2.5-7B}} & 
Base LLM (N=1) & 85.40 & — & 85.40 & — & — \\ 
& Zero-Shot CoT-SC (N=1) & 83.70 & 64.53 & 74.12 & 49.65 & 66.40 \\ 
& Zero-Shot AC-SC (N=1) & 84.85 & 64.96 & 74.91 & 52.71 & 67.04 \\ 
& SoftCoT-SC (N=1) & 85.81 & 72.44 & 79.13 & 60.61 & 67.52 \\ 
& LTA-Thinker (N=1) & \textbf{87.86} & \textbf{75.98} & \textbf{81.92} & \textbf{67.47} & \textbf{67.75} \\ 
\midrule
\specialrule{0.5pt}{1.0pt}{1.0pt}

\multirow{7}{*}{\makecell[l]{Qwem3-8B}} & 
Base LLM (N=1) & 89.84 & — & 89.84 & — & — \\ 
& Zero-Shot CoT-SC (N=10) & 92.22 & 76.77 & 84.50 & 70.96 & 84.56 \\ 
& Zero-Shot AC-SC (N=10) & 92.68 & 76.77 & 84.73 & 70.92 & 84.80 \\ 
& Coconut-SC (N=10) & 90.37 & 76.38 & 93.38 & — & — \\ 
& SoftCoT-SC (N=10) & 93.19 & 80.63 & 86.91 & 71.18 & 87.20 \\ 
& SoftCoT++ (N=10) & \textbf{93.65} & 84.09 & 88.87 & 71.22 & \textbf{88.16} \\ 
& LTA-Thinker (N=1) & \textbf{93.25} & \textbf{85.04} & \textbf{89.15} & \textbf{71.83} & \textbf{84.55} \\ 
\bottomrule
\end{tabular}
\caption{The main results table shows that LTA-Thinker achieves state-of-the-art (SOTA) performance on nearly all datasets. The "SC" in the table denotes self-consistency\cite{DBLP:conf/iclr/0002WSLCNCZ23} , a method where the model generates "N" outputs for a given question, and the most frequent answer among them is selected as the final result.}
\label{table1}
\end{table*}

\subsubsection{Distribution Scale Constraint}
Semantic alignment ensures relevance but does not differentiate between critical and non-critical information in the reasoning path. To learn to identify and focus on the most critical reasoning steps, we designed a contrastive learning-based Reasoning Focus Loss. The Reasoning Focus Loss aims to structurally expand the variance of the Latent Thought distribution. Its objective is to pull the question representation vector $\mathbf{e}_q$ closer to the critical reasoning steps (positive samples) in the representation space, while pushing it away from non-critical steps (negative samples). This calculation process is as follows:
\begin{itemize}
 \item \textbf{Sample Construction:} For each sample for training, we take the hidden state representations of the question from the output of the Latent Thought generation module as $\mathbf{e}_{\text{anchor}}$. The hidden state representations of each reasoning step in the golden truth chain of thoughts, $\{ \mathbf{s}_1, \dots, \mathbf{s}_M \}$, are treated as candidates. 
 \item \textbf{Dynamic Positive Sample Selection:}We calculate the cosine similarity between each candidate step $\mathbf{s}_j$ and the embedding representation of the final answer $\mathbf{e}_{\text{ans}}$. The step with the highest similarity is dynamically selected as the positive sample $\mathbf{s}_{\text{pos}}$, as it can be assumed to contribute the most to reaching the final answer. The remaining steps are considered negatives. Here we have removed the final step containing the answer.
 \item \textbf{Contrastive Loss Calculation:}We adopt a InfoNCE-like contrastive loss, using a temperature coefficient $\tau$ to adjust the similarity scores, aiming to maximize the similarity between the anchor and the positive sample while minimizing its similarity to all negative samples:
 \begin{equation}
 L_{\text{focus}} = - \log \frac{\exp(\text{sim}(\mathbf{e}_{\text{anchor}}, \mathbf{s}_{\text{pos}}) / \tau)}{\sum_{j=1}^{M} \exp(\text{sim}(\mathbf{e}_{\text{anchor}}, \mathbf{s}_j) / \tau)}
\end{equation}
where the temperature coefficient $\tau$ is set to 0.1. By this loss, Latent Thought distribution $Q_{\text{LTA-Thinker}} $ becomes a nonisotropic distribution. The model is motivated to understand the entire reasoning flow and to have its generated Latent Thought vectors guide the model to the critical nodes in the reasoning path.
\end{itemize}
Due to the effect of the Semantic Alignment Loss in the Distribution Locality constraint, the information contained in the positive sample can be transferred to the Latent Thought vectors via the question representation vector $\mathbf{e}_q$. The reason for not applying this contrastive loss directly to the Latent Thought vectors is that, as mentioned in Lemma 1, only when the magnitude of the perturbation vector $ \delta $  is sufficiently small can the perturbed sample $ L_{\text{Latent}} + \delta $ lie within the high probability density region of $ P_{\text{real}} $ . If applied directly to the Latent Thought vectors, it would cause the magnitude of the perturbation vector $ \delta $ to become too large, leading to $ \text{Var}[P_{\text{real}}]< \text{Var}[Q_{\text{LTA-Thinker}}]$.

Finally, the Reasoning Process. The input of the backbone LLM is augmented with Latent Thoughts, after being generated by a Transformer Block and constrained by three losses. All operations are performed in the latent space. The resulting vector is then fed into the Backbone LLM, guiding it to generate more precise and efficient reasoning paths. The Backbone LLM parameters are frozen.

\section{Experimental Results and Analyses}
\subsection{Datasets and Baselines}
We evaluated LTA-Thinker on multiple datasets covering reasoning. The datasets consists of three categories: mathematical, commonsense, and symbolic. For mathematical, we used the MATH-500~\cite{DBLP:conf/iclr/LightmanKBEBLLS24}, GSM8K~\cite{DBLP:journals/corr/abs-2110-14168}, and AQuA~\cite{DBLP:conf/acl/LingYDB17}. For commonsense, we used the StrategyQA~\cite{DBLP:journals/tacl/GevaKSKRB21}. For symbolic, we used the Date Understanding (DU) ~\cite{DBLP:journals/tmlr/SrivastavaRRSAF23} from Big-bench. We utilized the MATH-500, which is more widely used in the evaluation of LLM reasoning. Since the MATH-500 consists of 500 test samples selected from the full MATH dataset, we use the MATH dataset as the training set. In MATH-500, the answer to each test sample is in LaTeX format, which is highly inconvenient for evaluation. Therefore, during the actual evaluation process, we conduct further manual verification on samples that are incorrectly assessed by the machine. We selected 100 samples with relatively simple LaTeX formats that are easy to evaluate, thereby creating MATH-100.

For the baseline models being compared, we used the Qwen2.5-7B-Instruct and Qwen3-8B models. The compared baselines include: (1) Coconut~\cite{DBLP:journals/corr/abs-2412-06769}. (2) Zero-Shot CoT: This method uses the prompt proposed by ~\cite{DBLP:conf/iclr/SpragueYRJWSZYM25} and is evaluated in a zero-shot CoT manner. (3) Zero-Shot AC~\cite{DBLP:journals/corr/abs-2505-11484}: This method uses a smaller model from the same series as the baseline model to act as an auxiliary model. It is prompted to generate reasoning in discrete tokens, which is then applied to the CoT reasoning process of the Backbone LLM. (4) SoftCoT~\cite{DBLP:conf/acl/00010ZM25}. (5) SoftCoT++\cite{DBLP:journals/corr/abs-2505-11484}.

All models are trained on 4 NVIDIA A6000-48G GPU. Since LTA-Thinker employs randomly initialized Transformer block, we recommend setting the learning rate to 8e-5 or higher. Optimal results on nearly all datasets can be achieved within 10 epochs. We set the batch size to 16. The experimental results are the average of two results.

\subsection{The main experimental results and analysis}
The main experimental results are shown in Table \ref{table1} and Table \ref{table2}. It is important that LTA-Thinker achieved SOTA to all baseline methods in almost all datasets. Furthermore, in all experiments with the Qwen3-8B model, the baseline models used an "N=10" setting, meaning the model generated 10 reasoning chains for the same question and used the most frequent answer as the final answer. In contrast, LTA-Thinker, with only an "N=1" setting, surpassed the results of all baseline methods at "N=10". This simultaneously proves that the generated Latent Thought distribution from LTA-Thinker has a higher variance upper bound, indicating that this distribution can better approximate the golden truth distribution. Due to the excessive GPU memory consumption of SoftCoT++ during training, which is difficult to afford, we did not report its results on the MATH dataset.

\begin{table}[t]
\centering
\setlength{\heavyrulewidth}{1.2pt}
\setlength{\tabcolsep}{5pt} 
\small
\begin{tabular}{@{}llcc@{}}
\toprule
\multirow{2.5}{*}{\makecell[b]{Base LLM}} & 
\multirow{2.5}{*}{\makecell[b]{Baselines}} & 
\multicolumn{2}{c}{Challenge Math} \\
& & 
\makecell[b]{Math-100} & 
\makecell[b]{Math-500} \\ 
\midrule

\multirow{2}{*}{\makecell[l]{Qwen2.5-7B}} & 
SoftCoT-SC & 62.00 & \textbf{61.40} \\ 
& LTA-Thinker & \textbf{64.00} & 60.20 \\ 
\midrule
\specialrule{0.5pt}{1.0pt}{1.0pt}

\multirow{3}{*}{\makecell[l]{Qwem3-8B}} & 
Base LLM & 82.80 & 82.80 \\ 
& SoftCoT-SC & 79.00 & 80.80 \\ 
& LTA-Thinker & \textbf{88.00} & \textbf{84.60} \\ 
\bottomrule
\end{tabular}
\caption{Performance comparison on challenging MATH subsets (Math-100 and Math-500).}
\label{table2}
\end{table}

LTA-Thinker achieved corresponding SOTA results on both the Qwen2.5 and Qwen3 series models, demonstrating the applicability of our method. LTA-Thinker obtained excellent results across the three categories of mathematical, commonsense, and symbolic, proving the general applicability, effectiveness, and stability of our method. The results on the MATH-100 and MATH-500 show that our method is also effective in higher-difficulty complex reasoning problems. Due to the complexity of these problems, the performance improvement of LTA-Thinker on these two datasets is highly significant, mitigating the issue of diminishing marginal returns seen against the backdrop of high accuracy on other tasks. Regarding the results on the DU dataset, since it is a small-scale dataset and SoftCoT and SoftCoT++ evaluate it by loading parameters trained on other datasets—a process unknown to us—we only report the results using parameters trained for a single epoch on the DU dataset. We train for only one epoch on DU because the parameters in LTA-Thinker's Latent Thought generation module are randomly initialized and thus require a certain degree of adjustment. The results for the Base-LLM in the table were all obtained from publicly available sources online.

\begin{table}[t]
\centering
\label{tab:ablation_study}
\begin{tabular}{lc}
\toprule
\textbf{Experimental configuration} & \textbf{GSM8K} \\
\midrule
SFT-only                   & 89.61    \\
SFT + Con                  & 91.66    \\
SFT + KL                   & 91.21    \\
Liner-Assistant-SFT + Con + KL & 92.34    \\
SoftCoT-SFT + Con + KL     & 92.27    \\
SoftCoT-SC ($N=1$)         & 92.48    \\
SoftCoT++ ($N=1$)          & 92.48    \\
SoftCoT-SC ($N=10$)        & 93.19    \\
SoftCoT++ ($N=10$)         & \textbf{93.65}    \\
SoftCoT++ ($N=100$)        & \textbf{94.12}    \\
LTA-Thinker ($N=1$)        & \textbf{93.25}    \\
LTA-Thinker ($N=10$)       & \textbf{94.24}    \\
\bottomrule
\end{tabular}
\caption{Ablation experiments on the GSM8K dataset, where N denotes the number of inference responses.}
\label{table3}
\end{table}

\subsection{Ablation experiment results and analyses}
To evaluate LTA-Thinker more effectively, we conducted ablation studies as shown in Table \ref{table3}. Using LTA-Thinker (N=1) as the reference, "SFT-only" represents the result using only the SFT loss, while "SFT+Con" indicates the result after removing the Reasoning Focus Loss, and "SFT+KL" shows the result after removing the Semantic Alignment Loss. The "Linear-Assistant-SFT+Con+KL" variant represents the outcome of replacing the Transformer Block—originally a single-head self-attention layer plus a two-layer MLP—with a single linear layer. Furthermore, "SoftCoT-SFT+Con+KL" signifies the application of the multi-objective joint loss proposed in this paper to the SoftCoT method. In this context, "N" denotes the number of reasoning responses used in self-consistency.

Based on Table \ref{table3}, it can be concluded that: (1) Both the Semantic Alignment Loss and the Reasoning Focus Loss enhance the model's performance to a certain extent, and when both losses are removed, the model performs worst. (2) The result of "SoftCoT-SFT+Con+KL" being slightly lower than "Liner-Assistant-SFT+Con+KL" demonstrates the relationship between the model settings and the variance of the latent thought generation distribution discussed in this paper. Specifically, models that are not pre-trained and have a smaller parameter size will have larger upper and lower bounds on their variance. Due to excessive loss volatility and unfavorable for convergence, this result is the best result among multiple checkpoint evaluations with the same parameter settings as LTA-Thinker (N=1). (3) Due to the speciality of SoftCoT++, when N=1, the method is equivalent to SoftCoT-SC (N=1). (4) "LTA-Thinker (N=10)" achieves a better experimental result than "SoftCoT++ (N=100)", which proves the effectiveness of LTA-Thinker in parallel scaling and that LTA-Thinker has more efficient scaling feedback. In contrast, SoftCoT++ requires scaling to 100 responses to achieve optimal results. (5) "LTA-Thinker (N=10)" achieves the highest result, demonstrating that LTA-Thinker effectively utilizes the available information and is simple in structure and computationally efficient.

\begin{figure}[t]
  \centering
  \begin{subfigure}[b]{\columnwidth}
  \centering
    \includegraphics[width=0.98\textwidth]
    {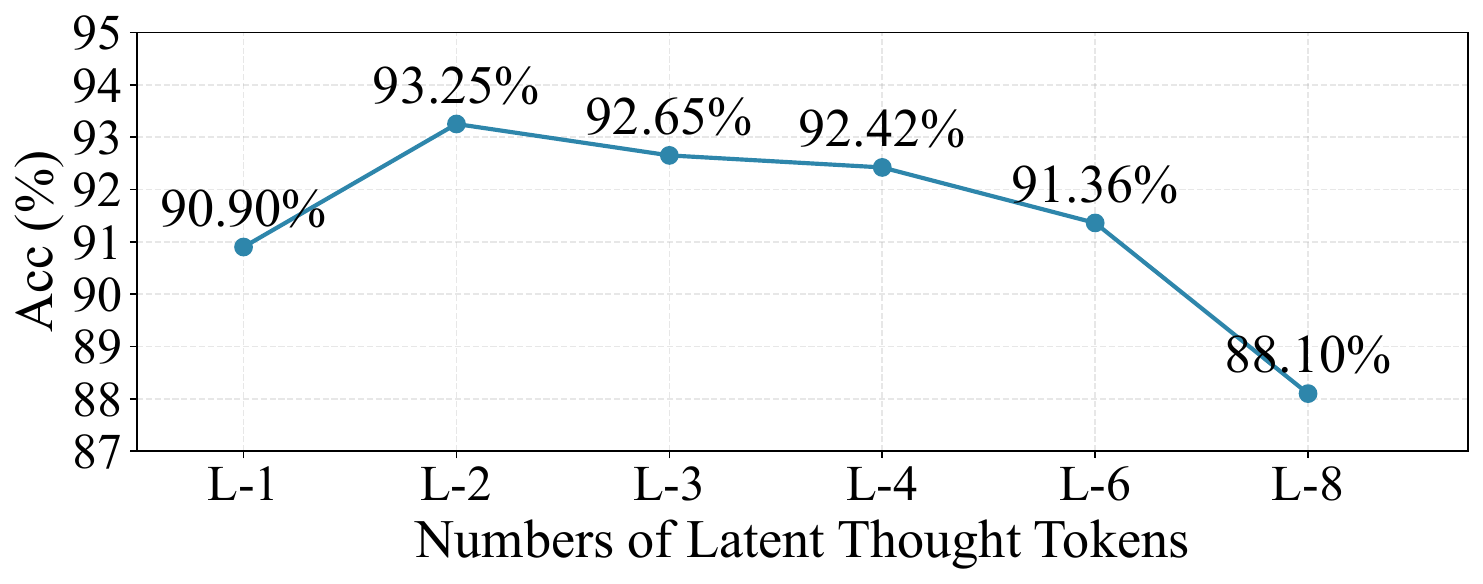}
    \captionsetup{justification=centering}
    \caption{Impact of Latent Thought tokens numbers}
    \label{fig:latent}
  \end{subfigure}
  
  \vspace{0.05cm} 
  
  \begin{subfigure}[b]{\columnwidth}
  \centering
    \includegraphics[width=0.98\textwidth]
    {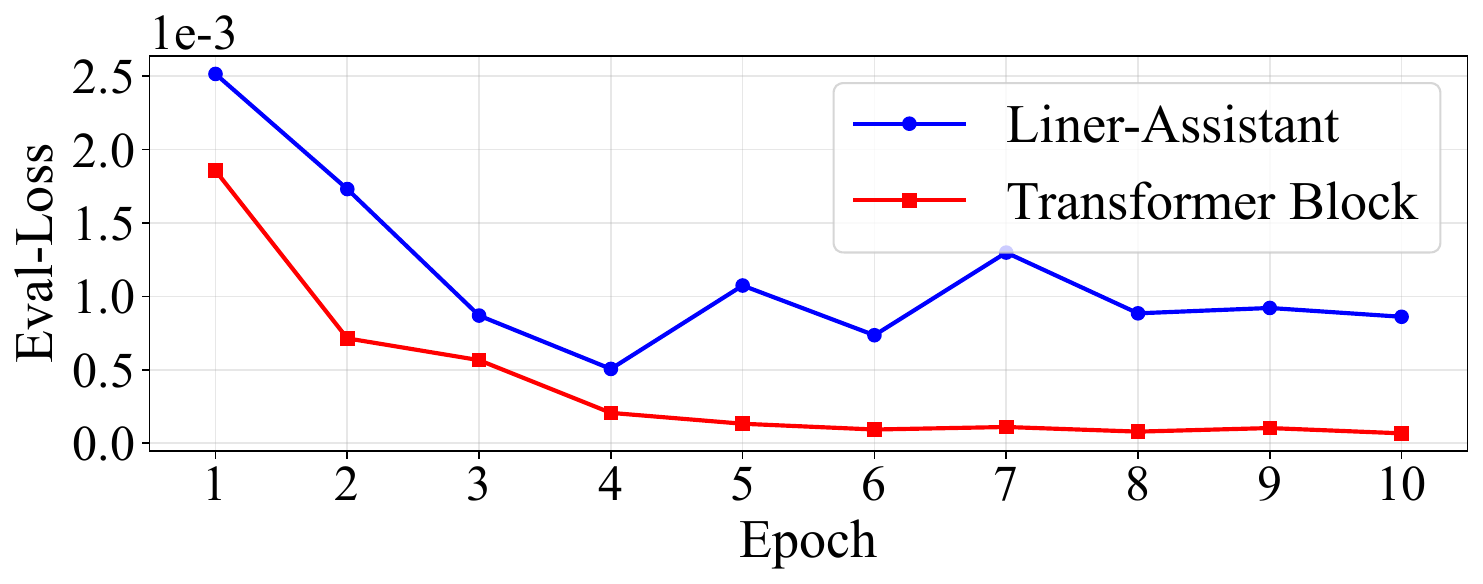}
    \captionsetup{justification=centering}
    \caption{Eval-Loss during training}
    \label{fig:eval}
  \end{subfigure}
  \caption{Curve Chart of Latent Thought Tokens number (Subfigure a above), and Models Convergence Effect Comparison Chart (Subfigure b below).}
  \label{fig:combined}
\end{figure}

To investigate the impact of the number of Latent Thought Tokens (L-N) on the model's results, we conducted experiments on the GSM8K, as shown in Figure \ref{fig:latent}. The experimental results indicate that the performance is optimal when L-N is 2, and as the number of L-N increases, the performance gradually decreases. The results for LTA-Thinker in both the main experiments and the ablation studies use L-N=2. To demonstrate that the convergence performance of the Transformer Block is superior to that of the Liner-Assistant, we plotted the loss curves for both experiments, as shown in Figure \ref{fig:eval}. This proves that the Liner-Assistant has poorer convergence; although it has a higher upper bound for variance, it exhibits extreme volatility during experimental tuning, which is not conducive to finding the optimal point. The losses were obtained with learning rate 1e-4 and the other parameters are all consistent.

\section{Conclusions}

In this paper, we propose LTA-Thinker, a framework using TTS to address LLM overthinking in complex reasoning tasks. The framework's core idea is to optimize the Latent Thought generated by the model to more closely approximate the ideal reasoning distribution. LTA-Thinker innovates in two areas: First, it utilizes a lightweight, randomly-initialized learnable prior generation architecture (a Transformer Block). This design can get closer to the golden truth distribution. Second, it introduces a distribution-based directional optimization paradigm that shapes the "direction" and "shape" of the Latent Thought by a multi-objective loss function. Experimental results demonstrate that LTA-Thinker achieves SOTA performance on multiple reasoning benchmarks. When N=1, it outperforms baselines such as SoftCoT ++ with N = 10, demonstrating its significant advantages in both reasoning efficiency and effectiveness.

\bibliography{aaai2026}
\clearpage

\lstset{
    basicstyle=\ttfamily\small,
    frame=single,
    breaklines=true,
    postbreak=\mbox{\textcolor{red}{$\hookrightarrow$}\space},
    escapechar=|,
    literate={{}{\{}}1 {{}}{{}}1 
}
\setcounter{secnumdepth}{2}
\section{Appendix}

\subsection{Proof process of Lemma 1}
\subsubsection{Lemma 1:}
If the magnitude of the perturbation vector $ \delta $ is sufficiently small, then the perturbed sample $ L_{\text{Latent}} + \delta $ will still be within a high-probability density region of $ P_{\text{real}} $.

Performing a Taylor expansion on the probability density function of 
$$
p(x + \delta) \approx p(x) + \nabla p(x)^\top \delta + \frac{1}{2} \delta^\top \nabla^2 p(x) \delta + \cdots
$$
When $ \|\delta\| \to 0 $ , the higher-order terms can be ignored, thus $ p(x + \delta) \approx p(x) $, meaning the perturbed sample still belongs to the high-probability region of the original distribution.
\subsection{Proof process of Lemma 2}
\subsubsection{Lemma 2:}
Let $P_{\text{real}}$ be the golden truth distribution (target distribution). Assume that the variance of $ P_{\text{real}} $ is $ \text{Var}[P_{\text{real}}]>0 $. Let $Q_1$ and $Q_2$ be two empirical distributions estimated from perturbed soft thoughts.

If the variances (covariance matrices) satisfy the condition $\text{Var}[Q_1] < \text{Var}[Q_2] \leq \text{Var}[P_{\text{real}}]$, then the KL divergence satisfies:
$$
\text{KL}(P_{\text{real}} \| Q_2) < \text{KL}(P_{\text{real}} \| Q_1)
$$
This inequality indicates that $ Q_2 $ provides a better approximation to the golden truth distribution $ P_{\text{real}} $ than $ Q_1 $.

\paragraph{Proof:}
Following the derivation in Appendix A.2, we assume $ P_{\text{real}}, Q_1, Q_2 $ are $d$-dimensional Gaussian distributions:
$$
P_{\text{real}} = \mathcal{N}(\mu, \Sigma), \quad Q_1 = \mathcal{N}(\hat{\mu}_1, \hat{\Sigma}_1), \quad Q_2 = \mathcal{N}(\hat{\mu}_2, \hat{\Sigma}_2)
$$
The condition on variances implies the positive definite ordering of covariance matrices:
$$
\hat{\Sigma}_1 < \hat{\Sigma}_2 \leq \Sigma
$$
The KL divergence between the target $P_{\text{real}}$ and an approximation $Q$ is given by:
\begin{equation}
\label{eq:kl_def}
\begin{split}
\text{KL}(P_{\text{real}} \| Q) = \frac{1}{2} \biggl[
&\text{tr}(\Sigma_Q^{-1} \Sigma) \\
+& (\hat{\mu}_Q - \mu)^\top \Sigma_Q^{-1} (\hat{\mu}_Q - \mu) \\
-& d + \log \frac{\det \Sigma_Q}{\det \Sigma}
\biggr]
\end{split}
\end{equation}
Based on the derivation in the paper, we assume that the expected values of the empirical means approximate the true mean, i.e., $\mathbb{E}[\hat{\mu}_1] \approx \mathbb{E}[\hat{\mu}_2] \approx \mathbb{E}[\mu]$. Consequently, the quadratic term (the second term in Eq. \ref{eq:kl_def}) approximates to 0 and can be ignored.

Let matrix $A = \Sigma_Q \Sigma^{-1}$. The equation simplifies to a function of matrix $A$:
\begin{equation}
\begin{split}
\text{KL}(P_{\text{real}} \| Q) &\approx \frac{1}{2} \left[ \text{tr}(A^{-1}) - d + \log \det A \right] \\
&= \frac{1}{2} (f(A) - d)
\end{split}
\end{equation}
where $f(A) = \text{tr}(A^{-1}) + \log \det A$. 

The function $f(A)$ achieves its global minimum when $A = I_d$ (the identity matrix), which implies $\Sigma_Q = \Sigma$. Since the function is convex around the minimum, the closer $\Sigma_Q$ is to $\Sigma$, the smaller the value of $f(A)$.

Given the condition $\hat{\Sigma}_1 < \hat{\Sigma}_2 \leq \Sigma$, the matrix $\hat{\Sigma}_2$ is closer to the optimal $\Sigma$ than $\hat{\Sigma}_1$ is. Therefore:

\begin{equation}
f(\hat{\Sigma}_2 \Sigma^{-1}) < f(\hat{\Sigma}_1 \Sigma^{-1})
\end{equation}
Substituting this back into the KL divergence formula, we conclude:
\begin{equation}
\text{KL}(P_{\text{real}} \| Q_2) < \text{KL}(P_{\text{real}} \| Q_1)
\end{equation}

\subsection{Datasets and Experiments}
All results in the experiment are accuracy results of the model on the corresponding dataset.

In the main experiment, the results of Base LLM on the MATH-100 dataset are approximately equivalent to those on MATH-500.

In all comparison experiments with SoftCoT and SoftCoT++, the same prompt or instruction was used for the experimental results on the same dataset.

Specifically, in all results using the Qwen2.5 model and Qwen3 model, we used the following similar prompt:
\maketitle

Here is the Python prompt template used for math problem solving:

\begin{Verbatim}[breaklines=true, breakanywhere=true, fontsize=\small]
input_template = (
    f"Solve the following math problem efficiently and clearly:\n"
    f"- For simple problems (2 steps or fewer):\nProvide a concise solution with minimal equation.\n"
    f"- For complex problems (3 steps or more):\n"
    f"Use this step-by-step format:\n\n"
    f"## Step 1: [Brief calculations]\n"
    f"## Step 2: [Brief calculations]\n"
    f"...\n"
    f"Regardless of the approach, always conclude with:\n"
    f"Therefore, the final answer is: $\\boxed{{answer}}$. I hope it is correct.\n"
    f"Where [answer] is just the final number or expression that solves the problem.\n\n"
    f"Problem: {question}"
)

if base_backbone in ["qwen3"] and split in ["train", "dev"]:
    input_template = f"Problem: {question}"
\end{Verbatim}

\vspace{1em}
Different prompts are used for training and evaluation, with the following conditions and formats:
\vspace{1em}

\begin{Verbatim}[breaklines=true, breakanywhere=true, fontsize=\small]
if base_backbone not in ["qwen3"] or split not in ["train", "dev"]:
    added_content = (
        f"Please think step by step and provide a detailed reasoning process."
        f"There are some prompts generated by a weaker assistant model. Some prompts maybe useful "
        f"while others maybe unuseful for your reasoning. "
        f"If the prompts are correct, you can use it as reference. "
        f"If the prompts are not correct, "
        f"you can ignore them and focus back to solving the problem.\n"
        f"Here are prompts: {soft_thoughts}"
    )
else:
    added_content = (
        f""
        f"Here are prompts from assistant model for reference: {soft_thoughts}"
    )
input_content += added_content
\end{Verbatim}

In the ablation experiments, “SFT-only” refers to the results obtained by using only the SFT loss; “SFT+Con” refers to the results obtained after removing the inference focus loss; and “SFT+KL” refers to the results obtained after removing the semantic alignment loss. These three experiments were implemented by setting the weight hyperparameters of the corresponding method losses to 0. Specifically, in the “SFT-only” experiment, let  $\lambda_{\text{sft}}=1$, $\lambda_{\text{align} }=0$, $\lambda_{\text{focus}}=0$. In the “SFT+Con” experiment, let  $\lambda_{\text{sft}}=0.5$, $\lambda_{\text{align} }=0.5$, $\lambda_{\text{focus}}=0$. In the “SFT+KL” experiment, let  $\lambda_{\text{sft}}=0.5$, $\lambda_{\text{align} }=0$, $\lambda_{\text{focus}}=0.5$.

For the results on the DU dataset, since we cannot determine which post-training checkpoint SoftCoT used for evaluation, we can only report a slightly lower accuracy rate.

Like SoftCoT and SoftCoT++, LTA-thinker essentially transforms the generation process of Latent Thought. The reason LTA-thinker and SoftCoT++ outperform SoftCoT is that they inject more information into Latent Thought. From a model structure perspective, SoftCoT++ outperforms SoftCoT because it trains the intermediate projection model more effectively, as evident from SoftCoT's training process. During training, SoftCoT expands the number of soft thought tokens to 32 and reduces it to 4 only during testing. This trick ensures the projection model is thoroughly trained. Similarly, our work noted this when following SoftCoT, so we began testing the injection of more information into Latent Thought, such as the two losses mentioned in this paper. The experimental results at the time were effective, with the best results on the GSM8k dataset exceeding 88 when using the Qwen2.5 model, which is encouraging. We also experimented with different initialization methods for Latent Thought, opting to use the reserved tokens \verb!<|extra_0|>!, \verb!<|extra_1|>!, etc. from the Qwen series models. However, the experimental results were not ideal. After analysis, we concluded that these reserved tokens had not undergone pre-training, resulting in limited information content. We then followed the work of SoftCoT++ and initialized Latent Thought with tokens present in the vocabulary. The experimental results showed a slight improvement. This result strengthened our confidence, leading us to refine the settings of the two losses and apply them to LTA-thinker.

For all datasets, we will submit the pre-processing, processing, and post-processing datasets together in the code. We will also provide the corresponding data processing scripts.

\subsection{Case Study}

\begin{Verbatim}[breaklines=true, breakanywhere=true, fontsize=\small]
{
  "question": "Janet\u2019s ducks lay 16 eggs per day. She eats three for breakfast every morning and bakes muffins for her friends every day with four. She sells the remainder at the farmers' market daily for $2 per fresh duck egg. How much in dollars does she make every day at the farmers' market?",
  "answer": "How many eggs does Janet sell? ** Janet sells 16 - 3 - 4 = <<16-3-4=9>>9 duck eggs a day.\nHow much does Janet make at the farmers' market? ** She makes 9 * 2 = $<<9*2=18>>18 every day at the farmer\u2019s market.\n#### 18"
}
\end{Verbatim}
The above is a sample of gsm8k data in JSON format. In the “answer” tag, “\texttt{\textbackslash n}” is used to separate the inference steps, and finally “\verb|\n####|” is used to separate the golden truth answer. The above text can be formatted as follows for easy viewing by readers.

\textbf{Problem Statement:} Janet's ducks lay 16 eggs per day. She eats three for breakfast every morning and bakes muffins for her friends daily using four eggs. She sells the remaining eggs at the farmers' market for \$2 per fresh duck egg. How much (in dollars) does she earn daily from the farmers' market?

\textbf{Solution Breakdown:}
\begin{enumerate}
    \item \textbf{Calculate daily eggs sold:}
    \[
    16\ \text{(total eggs)} - 3\ \text{(breakfast)} - 4\ \text{(baking)} = 9\ \text{eggs}
    \]
    
    \textbf{Compute daily revenue:}
    \[
    9\ \text{eggs} \times \$2\ \text{per egg} = \$18
    \]
\end{enumerate}
\textbf{Conclusion:} Janet earns \boxed{18} dollars daily from egg sales at the farmers' market.
\end{document}